\documentclass[runningheads]{llncs}

\usepackage{hyperref}
\usepackage[british]{isodate}
\usepackage[utf8]{inputenc}
\usepackage[utf8]{vietnam} 
\usepackage[english]{babel}
\usepackage{amsmath}
\usepackage{amssymb}
\usepackage{tikz}
\usepackage{pgfplots}
\usepackage{graphicx}
\usepackage{float}
\usepackage{bm}
\usepackage{caption}
\usepackage{subcaption}
\usepackage{booktabs}
\usepackage{verbatim}
\usepackage{cprotect}
\usepackage{multirow}
\usepackage{multicol}
\usepackage{fancyhdr}
\usepackage[font=itshape]{quoting}
\usepackage[backend=biber, style=numeric, sorting=none]{biblatex}
\usetikzlibrary{arrows, automata, fit, positioning, shapes, calc, decorations.pathreplacing, calligraphy}
\pgfplotsset{compat=1.18}

\definecolor{bluemark}{RGB}{100, 150, 235}
\definecolor{grey}{RGB}{220, 220, 220}
\definecolor{lightred}{RGB}{250, 200, 200}
\definecolor{lightgreen}{RGB}{200, 240, 210}
\definecolor{lightblue}{RGB}{200, 215, 250}
\definecolor{lightyellow}{RGB}{255, 245, 200}
\definecolor{lightgrey}{gray}{0.8}

\renewcommand{\rightmark}{Character Queries: Transformer-based Character Segmentation}

\makeatletter
\AtBeginDocument{
  \hypersetup{
    pdftitle={\@title},
    pdfauthor={Michael Jungo, Beat Wolf, Andrii Maksai, Claudiu Musat and Andreas Fischer}
  }
}
\makeatother

\title{Character Queries: A Transformer-based Approach to On-Line Handwritten Character Segmentation}
\titlerunning{Character Queries: Transformer-based Character Segmentation}

\author{%
  Michael Jungo\inst{1,2} \and
  Beat Wolf\inst{1} \and
  Andrii Maksai\inst{3} \and
  Claudiu Musat\inst{3} \and
  Andreas Fischer\inst{1,2}
}

\authorrunning{M. Jungo et al.}

\institute{%
  iCoSys Institute, University of Applied Sciences and Arts Western Switzerland \\
  \email{\{michael.jungo,beat.wolf,andreas.fischer\}@hefr.ch}
  \and
  DIVA Group, University of Fribourg, Switzerland \\
  \and
  Google Research \\
  \email{\{amaksai,cmusat\}@google.com}
}

\begin{document}

\maketitle

\begin{abstract}
  On-line handwritten character segmentation is often associated with handwriting recognition and even though
  recognition models include mechanisms to locate relevant positions during the recognition process, it is typically
  insufficient to produce a precise segmentation. Decoupling the segmentation from the recognition unlocks the potential
  to further utilize the result of the recognition. We specifically focus on the scenario where the transcription is
  known beforehand, in which case the character segmentation becomes an assignment problem between sampling points of
  the stylus trajectory and characters in the text. Inspired by the $k$-means clustering algorithm, we view it from the
  perspective of cluster assignment and present a Transformer-based architecture where each cluster is formed based on
  a learned character query in the Transformer decoder block. In order to assess the quality of our approach, we create
  character segmentation ground truths for two popular on-line handwriting datasets, IAM-OnDB and HANDS-VNOnDB, and
  evaluate multiple methods on them, demonstrating that our approach achieves the overall best results.
  
  \keywords{On-Line Handwriting \and Digital Ink \and Character Segmentation \and Transformer}
\end{abstract}

\thispagestyle{fancy}
\renewcommand{\headrulewidth}{0pt}
\lfoot{\small{\textit{Proceedings of the $17^{th}$ International Conference on Document Analysis and Recognition 2023 (ICDAR 2023)}}}

\section{Introduction}

\textbf{Relevance.} A significant advantage of using a stylus over a keyboard is its flexibility. As with pen and paper,
users can draw, write, link objects and make gestures like circling or underlining with ease \textendash{} all with
a handful of strokes. For digital ink to have a compelling value proposition however, many features associated with all
the use cases, that users have become accustomed to in an online environment, become relevant. They \textbf{go beyond
the initial act of writing} and cover layout and ink generative models like autocompletion and spelling correction. 

\textbf{Usefulness.} On-line handwriting character segmentation has as a goal the understanding of which parts of the
handwriting belong to which character. It complements handwriting recognition and enables functionalities like
generative modeling, particularly \textbf{spellchecking and correction}~\cite{inkorrect}, as well as ink-to-text
conversion and layout handling\cite{Jamboard,Freeform}.
What all these seemingly different tasks have in common is a need for character-level information. 

Character-level knowledge opens up the possibility for layout-preserving processing. For instance, when converting the
handwritten text into printed text, knowing the positions of the individual characters allow to generate printed text
that is precisely superimposed on top of the printed text, retaining the feeling of agency over the document (e.g.\ in
devices like Jamboard~\cite{Jamboard} and note-taking apps like FreeForm~\cite{Freeform}). Moreover, for education and
entertainment applications, knowing the positions of characters can unlock the capabilities such as animating individual
characters (e.g.\ in the Living Jiagu project the symbols of the Oracle Bone Script were animated as the animals they
represent~\cite{LivingJiagu}).

Individual character information is also important in handwriting generation models
~\cite{inkorrect,deepwriting,haines2016}. Examples include spellchecking and spelling correction. For spellchecking,
knowledge of word-level segmentation helps to inform the user about the word that was misspelled, e.g.\ marking the word
with a red underline, and the word-level segmentation is a natural byproduct of character-level segmentation. For
spelling correction, users could strike out a particular character or add a new one, and the remaining characters could
be edited such that the change is incorporated seamlessly, for example via handwriting generation
models~\cite{inkorrect}.

\textbf{Difficulty.} While the problem of character segmentation is fairly simple in case of printed text OCR, it is far
from being solved for handwriting \textendash{} both on-line and off-line. The problem is more difficult in settings
like \textbf{highly cursive scripts} (e.g.\ Indic) or simply cursive writing in scripts like Latin. Difficult cases
further include characters in Arabic script with ligatures, which vary in appearance depending on the surrounding of the
character and position in which they appear, and characters differing only by diacritics~\cite{hassan2019}.

In the academic on-line handwriting community, the progress on character segmentation is \textbf{limited by the absence
of the datasets with character segmentation}. Two notable exceptions are Deepwriting~\cite{deepwriting} and
BRUSH\cite{brush}.  The authors of Deepwriting  used a private tool to obtain a monotonic segmentation for the dataset.
This is limited, as it cannot accommodate cursive writing. In the BRUSH dataset an image segmentation model was used to
obtain the character segmentation. 

For this reason, most of the works in the on-line handwriting community \textbf{rely on synthetic datasets}. These are
created with either (1) segment-and-decode HMM-based approaches where character segmentation is a byproduct of
recognition~\cite{keysers}, or (2) hand-engineered script-specific features used in deep learning solutions, e.g.\ for
Indic and Arabic script~\cite{review_urdu,review_indic}, as well as mathematical expressions~\cite{review_math}. 

For off-line handwriting, character segmentation is usually based on the position of skip-token class spikes in the
Connectionist Temporal Classification (CTC)~\cite{ctc} logits \textendash{} which works well for images as the
segmentation is typically monotonic and separation of image by the spikes results in a reasonable segmentation (unlike
on-line handwriting, where due to cursive writing the segmentation is not monotonic).

Another difficulty that is associated with character segmentation is the \textbf{annotation}. Individually annotating
characters is hard and also time consuming. The most widely used handwriting datasets do not contain ground truth
information on individual characters. We can, however, infer a high quality synthetic ground truth using a time
consuming method that iteratively singles out the first character from the ink, based on temporal, spatial and stroke
boundary information. 

\textbf{Methods.} 
We compare multiple methods for character boundary prediction, with both a Long Short-Term Memory (LSTM)~\cite{lstm} and
Transformer~\cite{transformer} backbone and further comparing them with a simple $k$-means baseline. A first classifier
architecture, that accepts both an LSTM and a Transformer encoder, combines the individual point offsets with the CTC
spikes to determine which points represent character boundaries. This initial approach has a clear limitation in that it
is monotonic and cannot handle delayed strokes. We thus extend the Transformer classifier by including the character
information, where each character in the text becomes a query in the Transformer decoder block. To show its efficacy we focus
on the following approaches in an experimental evaluation on the publicly available IAM-OnDB and HANDS-VNOnDB datasets:
$k$-means, LSTM, Transformer for character boundary prediction and Transformer with
character queries.

The main \textbf{contributions} of this work can be summarized as follows:
\begin{itemize}
  \item We obtain character segmentation ground truths for the publicly available datasets IAM-OnDB and HANDS-VNOnDB
    from a high-quality approximation.
  \item We present a Transformer-based approach to the on-line handwritten character segmentation, where each expected
    output character is represented as a learned query in the Transformer decoder block, which is responsible for forming
    a cluster of points that belong to said character.
  \item We compare our approach to other methods on the IAM-OnDB and HANDS-VNOnDB datasets thanks to the newly
    obtained ground truth and demonstrate that it achieves the overall best results.
\end{itemize}
The newly created ground truths and the source code of our methods are publicly available at \href{https://github.com/jungomi/character-queries}{https://github.com/jungomi/character-queries}.

\section{Datasets}\label{sec:datasets}

\subsection{IAM On-Line Handwriting Database}\label{sec:iam}

IAM-OnDB~\cite{iamondb} is a database of on-line handwritten English text, which has been acquired on a whiteboard. It
contains unconstrained handwriting, meaning that it includes examples written in block letters as well as cursive
writing, and any mixture of the two, because it is not uncommon that they are combined in a way that is most natural to
the writer. With 221 different writers having contributed samples of their handwriting, the dataset contains a variety
of different writing styles.

\subsection{HANDS-VNOnDB}

The HANDS-VNOnDB~\cite{vnondb}, or VNOnDB in short, is a database of Vietnamese handwritten words. A characteristic of
Vietnamese writing, that is not found in English, is the use of diacritical marks, which can be placed above or below
various characters and even stacked. This poses an additional challenge, as the diacritics are often written with
delayed strokes, i.e.\ written after one or more characters have been written before finishing the initial character
containing the diacritics. Most notably in cursive and hasty writing, it is very common that the diacritics are
spatially displaced, for example hanging over the next character, which makes them disconnected in time as well as space
and therefore much more difficult to assign to the correct character.

\subsection{Ground Truth}

\begin{figure}[ht]
  \centering
  \begin{subfigure}[b]{\textwidth}
    \centering
    \begin{subfigure}[b]{\textwidth}
      \includegraphics[width=0.95\textwidth]{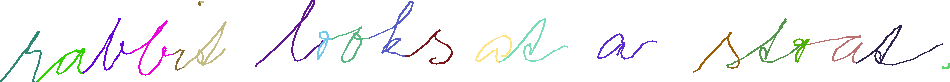}
      \caption*{rabbit looks at a stoat.}
    \end{subfigure}
    \begin{subfigure}[b]{\textwidth}
      \includegraphics[width=0.95\textwidth]{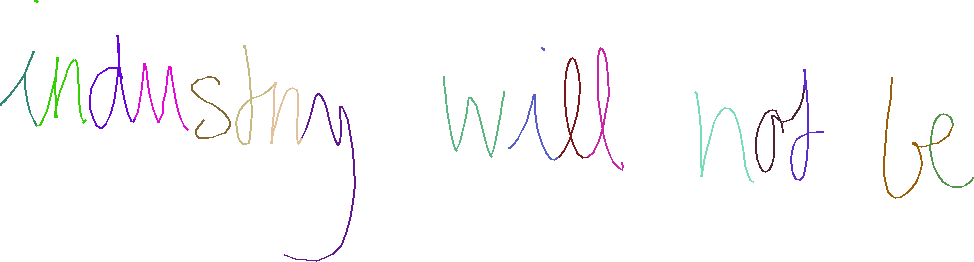}
      \caption*{industry will not be}
    \end{subfigure}
    \caption{IAM-OnDB Examples}
  \end{subfigure}
  \begin{subfigure}[b]{\textwidth}
    \centering
    \begin{subfigure}[b]{0.32\textwidth}
      \includegraphics[width=0.75\textwidth]{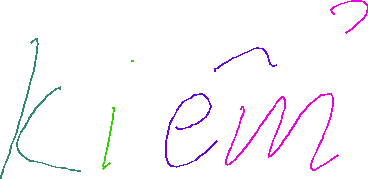}
      \caption*{kiểm}
    \end{subfigure}
    \begin{subfigure}[b]{0.32\textwidth}
      \includegraphics[width=0.75\textwidth]{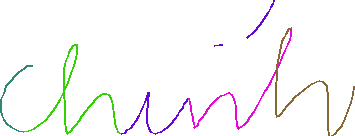}
      \caption*{chính}
    \end{subfigure}
    \begin{subfigure}[b]{0.32\textwidth}
      \includegraphics[width=0.75\textwidth]{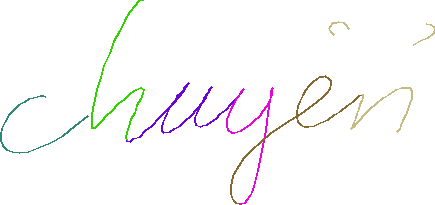}
      \caption*{chuyển}
    \end{subfigure}
    \begin{subfigure}[b]{0.32\textwidth}
      \includegraphics[width=0.75\textwidth]{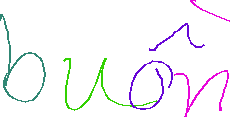}
      \caption*{buồn}
    \end{subfigure}
    \begin{subfigure}[b]{0.32\textwidth}
      \includegraphics[width=0.75\textwidth]{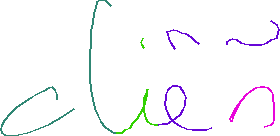}
      \caption*{diễn}
    \end{subfigure}
    \begin{subfigure}[b]{0.32\textwidth}
      \includegraphics[width=0.75\textwidth]{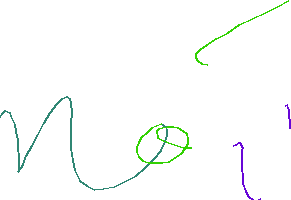}
      \caption*{nói}
    \end{subfigure}
    \caption{HANDS-VNOnDB Examples}
  \end{subfigure}
  \caption{
    Examples of the ground truth character segmentations that were obtained by iteratively separating the first
    character through an exhaustive search in regards to the character boundary. Each color represents a different
    character. Some imperfections are to be expected as it remains an approximation.
  }\label{fig:gt-samples}
\end{figure}

Since both of the publicly available datasets we are using do not have the ground truth character segmentation, we
resorted to obtaining a high-quality approximation of it (similar approach was applied, for example,
by~\cite{kotani2020} where an image-based approach was used for obtaining the ground truth approximation). To obtain it,
we repeatedly separated the first character from the rest of the ink, by performing an exhaustive search for the
character boundary with potential cuts based on temporal information, spatial information, and stroke boundaries, and
with the best candidate selected based on the likelihood that the first character indeed represents the first character
of the label, and the rest matches the rest of the label, with likelihood provided by a state-of-the-art recognizer
model~\cite{carbune}.
Such an approach is clearly not feasible in a practical setting due to the high computational cost, but allowed us to
produce a high quality ground truth approximation, from which any of the models described below could be trained.
\autoref{fig:gt-samples} illustrates some ground truth examples from the IAM-OnDB and the HANDS-VNOnDB. Despite the high
quality of the ground truth it remains an approximation and therefore some small imperfections are present, as evidenced
by some of the examples.

\section{Methods}

\subsection{K-Means}

For the initial baseline system, we chose to use a $k$-means~\cite{k-means} based approach. To segment the handwriting,
the points are clustered into $k$ different clusters, where $k$ is equal to the number of characters present in the
already known text. Two methods have been implemented for the initialization of the centroids, the standard random
implementation, and a second implementation which uses the points where the CTC spikes occurred as the initial
centroids. Clustering is mainly based on the geometric locations of the points but the stroke information was also
included, as it still adds value for points that are not clearly separable purely based on their position. Since the
horizontal position is much more indicative of the character it might belong to, the $x$ coordinate was weighted much
stronger than the $y$ coordinate. This heavily relies on the horizontal alignment of the writing and causes an inherent
limitation for cases where the alignment deviates from the ideal representation, e.g. for strongly slanted handwriting.

\subsection{Character Boundary Prediction with LSTM / Transformer}\label{sec:lstm}

The input of the model for the character boundary prediction is a sequence of sampling points and the output is
a classification of whether a point is a character boundary or not. Intuitively, an LSTM~\cite{lstm} can be employed for
this, as it is particularly well suited to work with a sequence based representation. Given that the output remains
a sequence but is not required to recognize which character it is, it is sufficient to have the tokens \texttt{<start>},
\texttt{<char>} and \texttt{<none>}, which signify the start of the character (boundary), being part of the current
character and not belonging to a character at all, respectively. An important note about the \texttt{<none>} token is,
that there is no point in the available ground truth that does not belong to any character, simply because of the
exhaustive nature of the ground truth creation, as a consequence it is repurposed to indicate that the point does not
belong to the current character, which primarily refers to delayed strokes. Due to the lack of back references in this
approach, it will just be considered as not part of any character.

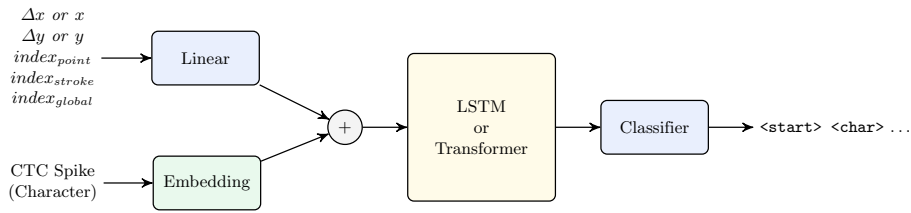
\begin{figure}[ht]
  \centering
  \scalebox{1.19}{
  \begin{tikzpicture}[>=stealth', auto, align=center, node distance=0.5cm, every node/.style={scale=0.6},
    cluster/.style={circle, text=white, fill=lightblue, minimum size=0.8cm, node distance=0.2cm},
    operator/.style={circle, draw, fill=lightgrey, fill opacity=0.2, text opacity=1},
    nn/.style={rectangle, draw, minimum width=2cm, minimum height=1cm, rounded corners=2},
    linear/.style={nn, fill=lightblue, fill opacity=0.4, text opacity=1},
    embedding/.style={nn, fill=lightgreen, fill opacity=0.4, text opacity=1},
    transformer/.style={nn, fill=lightyellow, fill opacity=0.4, text opacity=1, minimum size=2.75cm},
    ]
    \node[] (input) {$\Delta{x}$ \textit{or} $x$\\$\Delta{y}$ \textit{or} $y$\\$index_{point}$\\$index_{stroke}$\\$index_{global}$};
    \node[] (ctc) [below=of input] {CTC Spike\\(Character)};
    \node[embedding] (embedding) [right=0.55cm of ctc] {Embedding};
    \node[linear] (proj) [right=0.56cm of input] {Linear};
    \node[operator] (add) [above right=0.18cm and 0.8cm of embedding] {$+$};
    \node[transformer] (lstm) [right=of add] {LSTM\\or\\Transformer};
    \node[linear] (classifier) [right=of lstm] {Classifier};
    \node (output) [right=of classifier] {\texttt{<start> <char>} \dots{}};

    \path[draw,->]
      (input) edge node {} (proj)
      (ctc) edge node {} (embedding)
      (proj) edge node {} (add)
      (embedding) edge node {} (add)
      (add) edge node {} (lstm)
      (lstm) edge node {} (classifier)
      (classifier) edge node {} (output)
    ;
  \end{tikzpicture}
  }
  \caption{
    \textbf{Architecture of the boundary prediction model.} 
    A feature vector is created from the $x,y$-coordinates ($\Delta{x},\Delta{y}$ for the LSTM and absolute coordinates
    for the Transformer) and the stroke information, where the pair of $index_{stroke}$ and $index_{point}$ indicate
    which stroke the point belongs to and which point it is within the stroke, as well as the global position
    with $index_{global}$. For points where a CTC spike occurred, an embedding vector of the identified character is
    added to the existing feature vector. The resulting feature vector is processed either by bidirectional LSTMs or
    a Transformer and followed by a linear classifier to produce the boundary prediction.
  }\label{fig:lstm-arch}
\end{figure}

\autoref{fig:lstm-arch} outlines the architecture of the boundary prediction model. A sequence of points is given as the
input, where each point contains the $x$ and $y$ coordinates, as well as information at which part of the sequence it
occurred. Handwriting is almost always performed in multiple strokes, which can be a helpful indicator of where
a character might begin, therefore it is conveyed to the model with the pair of $index_{stroke}$ and $index_{point}$ to
identify the stroke it belongs to and which point it is within the stroke, while $index_{global}$ is also provided in
order facilitate locating the point globally. All this information is transformed by a linear layer to create a higher
dimensionality feature vector that is more appropriate for the LSTM. Additionally, for each point where a CTC spike
occurs, an embedding vector is created from the character it corresponds to, and added to the existing feature vector.
Afterwards, the feature vector is processed by multiple bidirectional LSTMs followed by a linear classifier to produce
the boundary predictions.

Transformers~\cite{transformer} are also widely used in sequence based task and are generally highly successful in many
situations where LSTMs perform well, hence the same architecture can be used with a Transformer instead of an LSTM. Some
minor changes to the input are required compared to the LSTM. The $x,y$-coordinates were given as deltas
($\Delta{x},\Delta{y}$) in regards to the previous point due to the recurrent nature of the LSTM, which turned out to
work slightly better than the absolute coordinates. Since Transformers do not have any recurrence, there is no reference
point for the deltas, therefore the absolute coordinates are the only viable option. Even though $index_{global}$ might
be considered to be more important for Transformers, it is not used because the same effect is achieved by
the positional encoding that is added to the Transformer to explicitly handle the positional information.

\subsubsection{Post-Processing}\label{sec:lstm-post-process}

In order to assign the points to the respective characters, the sequence of tokens needs to be processed such that the
point with the \texttt{<start>} token and all points marked as \texttt{<char>} up to the next \texttt{<start>} token are
assigned to the expected output characters from left to right. Technically, the model is not limited to produce exactly
number of expected characters, but is supposed to learn it. Unfortunately, it does occur that too many characters are
predicted, hence we additionally restrict the output to the desired number of characters by removing the segments with
the smallest number of points, as we are specifically interested in assigning the points to the expected characters.
This is a limitation of this particular design for the character boundary.

\subsection{Transformer with Character Queries}

Given that the design of the character boundary prediction in \autoref{sec:lstm} revolves around sequences of points, it
is impossible to handle delayed strokes appropriately. Furthermore, the expected output characters are not integrated
into the model, even though they are at least represented in the input features by the CTC spikes. This not only
necessitates some post-processing, due to possible oversegmentation,
but also eliminates any potential for the model to adapt to a specific character. To address these shortcomings, we
design a Transformer-based architecture that integrates the expected output characters into the Transformer decoder block by
using them as queries.

\subsubsection{Related Work}

In recent years, Transformers have been applied to many different tasks in various domains. With the necessity to adapt
to domains not initially suited for Transformers, due to the inherently different structures compared to the familiar
sequence based tasks, new Transformer-based approaches have been developed. In particular, in the domain of Computer
Vision (CV), a lot of creative designs have emerged~\cite{vision-survey-1,vision-survey-2}. One of these novel
approaches was pioneered by the DEtection TRansformer (DETR)~\cite{detr}, called object queries, where each learned
query of the Transformer decoder block represents one object that has to be detected. Later on the query based approach has
found its way to image segmentation tasks~\cite{max-deeplab,query-instances,mask2former}.

Only recently, the $k$-means Mask Transformer~\cite{k-means-transformer} introduced a Transformer-based approach that
was inspired by the $k$-means clustering algorithm, where the authors discovered that updating the object queries in the
cross-attention of the Transformer decoder block was strikingly similar to the $k$-means clustering procedure. While their
approach was made for image based segmentation, it can easily be adapted to our task of on-line handwritten
segmentation, which happens to remove a lot of the complexity that is only needed for images, mostly due to downsampling
of the image and upsampling of the segmentation masks. Inspired by their findings and the fact that our baseline
algorithm has been $k$-means, we design a Transformer-based architecture where the queries represent the characters that
should be segmented.

\subsubsection{Overview}

For each character that needs to be segmented, a query in the Transformer decoder block is initialized with the embedding of that
particular character and a positional encoding, which is necessary to distinguish two or more instances of the same
character. In that regard, the character embedding provides the information to the model as to which particular patterns
to pay attention to, while the positional encoding is primarily used to ensure that the order of the characters is
respected. Having the available characters tightly integrated into the model, eliminates the post-processing
completely, which is due to the fact that the characters were created from a sequence in the boundary prediction models,
whereas now the points are simply assigned to the respective characters, reminiscent of clustering algorithms such as
$k$-means. Additionally, it opens up the possibility to handle delayed strokes correctly without any modification as
long as they are represented adequately in the training data.

\subsubsection{Architecture}

\begin{figure}[ht]
  \definecolor{cluster1_colour}{HTML}{308774}
  \definecolor{cluster2_colour}{HTML}{32d102}
  \definecolor{cluster3_colour}{HTML}{e805da}
  \definecolor{cluster4_colour}{HTML}{846a32}
  \definecolor{annotations}{HTML}{8b8b8b}
  \centering
  \scalebox{1.148}{
  \begin{tikzpicture}[
      >=stealth',
      auto,
      align=center,
      node distance=0.5cm,
      every node/.style={scale=0.6},
      cluster/.style={circle, text=white, fill=lightblue, minimum size=0.8cm, node distance=0.2cm},
      operator/.style={circle, draw, fill=lightgrey, fill opacity=0.2, text opacity=1},
      nn/.style={rectangle, draw, minimum width=1.8cm, minimum height=0.9cm, rounded corners=2},
      linear/.style={nn, fill=lightblue, fill opacity=0.4, text opacity=1},
      embedding/.style={nn, fill=lightgreen, fill opacity=0.4, text opacity=1},
      transformer/.style={nn, fill=lightyellow, fill opacity=0.4, text opacity=1, minimum size=2.75cm},
    ]
    \node[] (input) {$x$\\$y$\\$index_{point}$\\$index_{stroke}$};
    \node[] (ctc) [below=of input] {CTC Spike\\(Character)};
    \node[embedding] (embedding) [right=of ctc] {Embedding};
    \node[linear] (proj) [right=0.56cm of input] {Linear};
    \node[operator] (add) [above right=0.18cm and 0.8cm of embedding] {$+$};
    \node[transformer] (encoder) [right=0.8cm of add] {Transformer\\Encoder};
    \node[transformer] (decoder) [right=1.2cm of encoder] {Transformer\\Decoder};
    \node (encoder_x) [below left=0cm and 0cm of encoder.north east] {\textbf{\textcolor{annotations}{$\times N$}}};
    \node (decoder_x) [below left=0cm and 0cm of decoder.north east] {\textbf{\textcolor{annotations}{$\times M$}}};
    \coordinate (enc_split) at ([xshift=0.6cm]encoder.east);
    \node[linear] (classifier_enc) [above=1.33cm of enc_split] {Linear};
    \node[linear] (classifier_dec) [above=of decoder] {Linear};
    \node[operator] (matmul) [above=of classifier_dec] {$\times$};

    \node[embedding] (cluster_embedding) [below=of decoder] {Embedding\\\textit{Character + Position}};
    \node[] (cluster_centre) [below=of cluster_embedding] {};
    \node[cluster, fill=cluster1_colour] (cluster1) [below left=0.1cm and 0.72cm of cluster_centre] {t};
    \node[cluster, fill=cluster2_colour] (cluster2) [right=of cluster1] {o};
    \node[cluster, fill=cluster3_colour] (cluster3) [right=of cluster2] {d};
    \node[cluster, fill=cluster4_colour] (cluster4) [right=of cluster3] {o};

    \node (input_points) [above right=0.2cm and 0cm of input.north west] {\includegraphics[scale=3.0]{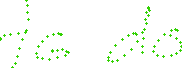}};
    \node (output) [above=of matmul] {\includegraphics[scale=3.0]{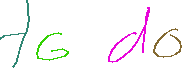}};

    \node (dimensions) [below right=1.0cm and 0.0cm of ctc.south west, align=left] {
      \small{\textcolor{annotations}{\textbf{\underline{Dimensions}}}}\\
      \small{\textcolor{annotations}{$p =$ Number of points}}\\
      \small{\textcolor{annotations}{$c =$ Number of characters/clusters}}\\
      \small{\textcolor{annotations}{$d_{h} =$ Size of hidden dimension}}\\
      \small{\textcolor{annotations}{$d_{f} =$ Size of final hidden dimension}}
    };

    \path[draw,->]
      (input) edge node {\small{\textcolor{annotations}{$p \times 4$}}} (proj)
      (ctc) edge node {\small{\textcolor{annotations}{$p$}}} (embedding)
      (proj) edge node {\small{\textcolor{annotations}{$p \times d_{h}$}}} (add)
      (embedding) edge node [below] {\small{\textcolor{annotations}{$p \times d_{h}$}}} (add)
      (add) edge node {\small{\textcolor{annotations}{$p \times d_{h}$}}} (encoder)
      (encoder) edge node [below] {\small{\textcolor{annotations}{$p \times d_{h}$}}} (decoder)
      (cluster_centre) edge node {\small{\textcolor{annotations}{$c$}}} (cluster_embedding)
      (cluster_embedding) edge node {\small{\textcolor{annotations}{$c \times d_{h}$}}} (decoder)
      (enc_split) edge node {} (classifier_enc)
      (decoder) edge node {\small{\textcolor{annotations}{$c \times d_{h}$}}} (classifier_dec)
      (classifier_enc) |- node [align=center, pos=0.75] {\small{\textcolor{annotations}{$p \times d_{f}$}}} (matmul)  
      (classifier_dec) edge node {\small{\textcolor{annotations}{$c \times d_{f}$}}} (matmul)
      (matmul) edge node {\small{\textcolor{annotations}{$p \times c$}}} (output)
    ;

    \draw[thick, decorate, decoration={brace, mirror}] ($(embedding.south west)+(-0.5,-0.275)$) -- node [below=0.15cm] {Features} ($(add.south east)+(0.0,-1.0)$);
    \draw[thick, decorate, decoration={brace, mirror}] ($(classifier_dec.south east)+(0.25,0.0)$) -- node [right=0.15cm] {Classification} ($(matmul.north east)+(0.65,0.0)$);

  \end{tikzpicture}
  }
  \caption{
    \textbf{Architecture of the Character Query Transformer.}
    Like the boundary prediction models, the feature vector is created from the $x,y$-coordinates, the stroke
    information and the CTC spikes. A Transformer encoder takes the feature vector and creates an encoded vector, which
    is given to the Transformer decoder block in combination with an embedding of the desired characters to be segmented
    (including their position within the text). The classification is achieved by a matrix multiplication between the
    output of the encoder and the decoder, after a linear transformation of each of the respective outputs, in order to
    assign each point to one character.
}\label{fig:transformer-arch}
\end{figure}

This model will subsequently be referred to as the Character Query Transformer and its architecture is outlined in
\autoref{fig:transformer-arch}. The input features remain the same as for the boundary prediction model, where the
feature vector is created from the $x,y$-coordinates, the stroke information through the pair of $index_{stroke}$ and
$index_{point}$, and the CTC spikes with an embedding of the identified character. A Transformer encoder is applied to
the feature vector to create a new encoded vector, $E \in \mathbb{R}^{p \times d_{h}}$, that captures more pertinent
information by virtue of the self-attention which incorporates the relation between the points. Afterwards,
a Transformer decoder block takes the encoded vector in combination with the character queries, which are created from the
expected output characters by applying a learned character embedding and positional embedding based on their position
within the text.

The output of the decoder, $D \in \mathbb{R}^{c \times d_{h}}$, cannot be used directly to create a classification for
each point, as it is merely a latent representation of the clusters, hence it has the dimensions $c \times d_{h}$, where
$c$ is the number of characters and $d_{h}$ the size of the hidden dimension. While the points have been assimilated
into $D$ through the cross-attention in the decoder, the exact association between points and characters must be done
with an additional step. This can be achieved with $E D^{T} \in \mathbb{R}^{p \times c}$, a matrix multiplication
between $E$ and $D$, the outputs of the encoder and decoder respectively. Normally, a classifier would be applied
afterwards, but because the dimensions of $p \times c$ are dynamic, since both $p$ (number of points) and $c$ (number of
characters) vary depending on the input, that is not possible. As an alternative, a linear transformation is applied
separately to $E$ and $D$ before the matrix multiplication.

\subsubsection{Positional Encoding of the Character Queries}

Transformers do not inherently have any sense of position of the inputs, as they do not contain any recurrence or
convolutions, which implicitly take the order into account. To alleviate this issue, a separate positional encoding is
added to the input to explicitly encode the positional information into the input. It is most commonly achieved with
a sinusoidal positional encoding, where the sine and cosine functions are used with different frequencies:

\begin{equation}
  \begin{split}
    PE_{(pos, 2i)} = \sin(\frac{pos}{10000^{2i/d}}) \\
    PE_{(pos, 2i + 1)} = \cos(\frac{pos}{10000^{2i/d}})
  \end{split}
\end{equation}

\begin{figure}[ht]
  \centering
  \begin{tikzpicture}
    \begin{axis}[
        ybar=-2.5pt,
        bar width=2.5pt,
        legend style={draw=none, font=\small},
        legend image code/.code={
          \draw [#1] (0cm,-0.1cm) rectangle (0.2cm,0.25cm);
        },
        xlabel={Position},
        ylabel={Normalized Mean Values},
        xlabel near ticks,
        ylabel near ticks,
        xtick pos=left,
        ytick pos=left,
        every x tick/.style={color=black, thin},
        every y tick/.style={color=black, thin},
        tick align=outside,
        axis on top,
        width=0.9\textwidth,
        height=0.55\textwidth,
        xmin=-3,
        xmax=103,
    ]
      \addplot+[restrict x to domain=0:100] table[x=pos, y=sincos, col sep=comma] {pos-data.csv};
      \addplot+[restrict x to domain=0:100] table[x=pos, y=learned, col sep=comma] {pos-data.csv};
      \legend{Sinusoidal, Learned}
    \end{axis}
  \end{tikzpicture}
  \caption{
    \textbf{Positional Encoding for Character Queries.}
    For each position the normalized mean values of the vector in the positional encoding are displayed. The sinusoidal
    encoding (\textcolor{blue}{blue}) follows a smooth trend with small variations between the positions, whereas the
    learned encoding (\textcolor{red}{red}) exhibits much larger differences between positions, which makes the
    distinction between multiple instances of the same character much more apparent to the model.
  }\label{fig:pos-enc}
\end{figure}
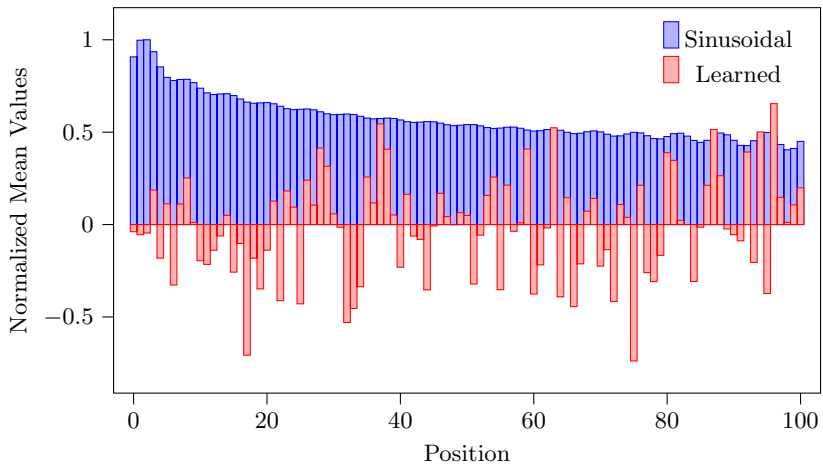

Due to the characteristics of the sinusoidal functions the transition between the positions remains predictable and
smooth, therefore the input features are not disrupted but rather slightly enhanced. Generally, this is a desirable
property, but in the case of the character queries, a more recognizable distinction between the position is needed,
because multiple instances of the same character need to be treated as completely separate. For this purpose, a learned
positional encoding is used instead. \autoref{fig:pos-enc} depicts the normalized mean values of the vector at each
position in the positional encoding for the sinusoidal (\textcolor{blue}{blue}) and learned encodings
(\textcolor{red}{red}) respectively. In the learned encoding it is clearly visible that there are a lot more large
differences between two positions, indicating that a clear distinction between them does benefit the model and its
capabilities to distinguish between multiple instances of the same character. On the other hand, the sinusoidal encoding
keeps a smooth transition between the positions and therefore lacks the clear distinguishing aspect, and in our
experience it was simply not enough to separate multiple instances of the same character.

\section{Experiments}

In this section we evaluate the four methods, namely the $k$-means, LSTM, Transformer (character boundary prediction)
and Character Query Transformer on the IAM-OnDB and the HANDS-VNOnDB, as well as combining the two dataset to see whether more
data with slightly different characteristics are beneficial to the overall results. And finally, an ablation study on
the usefulness of the CTC spikes is conducted. All results are evaluated based on the mean
Intersection over Union (mIoU) between the points in the segmented characters.

\subsection{Setup}

The experiments for the $k$-means are performed with Scikit-Learn~\cite{scikit-learn} whereas all other methods are
implemented in PyTorch~\cite{pytorch}. For the $k$-means the weights of the input features are set to 1 for the
$x$-coordinate, 0.04 for the $y$-coordinate and 224 for the stroke information. All PyTorch models use a dimension of
256 for all layers, i.e.\ embedding dimension, hidden dimension of LSTM/Transformer and the final hidden dimension, as
well as a dropout probability of 0.2. The LSTM consist of 3 bidirectional layers with the Rectified Linear Unit
(ReLU)~\cite{relu} as an activation function, whereas the Transformer uses 3 layers per encoder and decoder with
8 attention heads and Gaussian Error Linear Unit (GELU)~\cite{gelu} instead of ReLU. Label smoothing with $\epsilon
= 0.1$~\cite{label-smoothing} is employed in the cross-entropy loss function. AdamW~\cite{adamw} is used as an optimizer
with a weight decay of $10^{-4}$, $\beta_1 = 0.9$ and $\beta_2 = 0.98$. The learning rate is warmed up over 4\,000 steps
by increasing it linearly to reach a peak learning rate of $3 \cdot 10^{-3}$ for the LSTM and $10^{-3}$ for the
Transformer. Thereafter, it is decayed by the inverse square root of the number of steps, following the learning rate
schedule proposed in~\cite{transformer}. Additionally, Exponential Moving Average (EMA)~\cite{ema} is applied during the
training to obtain the final weights of the model.

\subsection{Results}

\subsubsection{IAM-OnDB}

On the IAM-OnDB the $k$-means baseline already achieves very respectable results of up to 91.05\% mIoU (\autoref{tab:results-iam}). Considering that
it is the simplest of the methods and does not need any training beforehand. The high mIoUs can be attributed to the
fact that in the English language most characters do not have the potential of creating much overlap with the next one,
unless the cursive writing is slanted excessively. As $k$-means relies heavily on the spatial position, it is capable of
separating a majority of the cases, particularly on block letters. 
The LSTM is the strongest on this dataset with an mIoU of \textbf{93.72\%} on the \textit{Test Set F}, suggesting that the model is expressive enough to handle more
difficult cases. A known limitation is that it cannot handle the delayed strokes but as only a small number of points
are actually part of the delayed strokes, in addition to delayed strokes being fairly rare in the first place, the
overall impact on the mean Intersection over Union (mIoU) is rather small.
On the other hand, the Transformer does not reach the same level of accuracy, and is even slightly below the $k$-means.
This is presumably due to limited amount of data, which does not satisfy the need of the generally data intensive
Transformer models.
Similarly, the Transformer with the character queries cannot establish the quality of results that is demonstrated in
other experiments. Having an mIoU that is roughly 3\% lower than the Transformer for the character boundary detection, is most
likely because of the character queries being learned, and the same data limitation applies to it, hence it cannot reach its full potential.

\begin{table}[ht]
  \centering
  \caption{
    \textbf{IAM-OnDB results.} All models were trained using only the IAM-OnDB training set
    and the best model was determined by the mIoU on the validation set.
  }\label{tab:results-iam}
  \begin{tabular}{c|cc}
    \toprule
    Model & \textit{Test Set T} & \textit{Test Set F} \\
    \midrule
    K-Means & 88.94 & 91.05 \\
    LSTM & \textbf{89.55} & \textbf{93.72} \\
    Transformer & 86.18 & 90.34 \\
    Character Query Transformer & 83.53 & 87.48 \\
    \bottomrule
  \end{tabular}
  \caption*{mean Intersection over Union (mIoU)}
\end{table}

\subsubsection{HANDS-VNOnDB}

The results for the HANDS-VNOnDB in \autoref{tab:results-vnondb} paint a very different picture from the IAM-OnDB results. The
$k$-means does not reach quite the same level of accuracy on the Vietnamese characters as on English characters, which
is mainly related to the additional complexity of Vietnamese characters, such as the use of diacritics, which can very
easily shift in such a way that it might be considered as part of another character when focusing only on the spatial as
well as temporal location of the points. This is a prime example, where additional language information is needed to
accurately segment such characters.
A much bigger difference to the previous results is observed in the LSTM, which is significantly worse than any other
method with an mIoU of just under 50\%. One expected reason for the degradation is the much more prominent use of delayed
strokes. In this situation, the LSTM exhibits significant problems to accurately predict the character boundaries. The
Transformer model performs better but only achieves a similar performance as the k-means baseline.
By far the best results are
achieved by the Character Query Transformer with a staggering \textbf{92.53\%} mIoU on the \textit{Test Set}, which is over 13\%
better than the next best method. This demonstrates that the approach is in fact working as the delayed strokes
are no longer an inherent limitation.

\begin{table}[ht]
  \centering
  \caption{
    \textbf{HANDS-VNOnDB results.} All models were trained using only the HANDS-VNOnDB training set
    and the best model was determined by the mIoU on the validation set.
  }\label{tab:results-vnondb}
  \begin{tabular}{c|c}
    \toprule
    Model & \textit{Test Set} \\
    \midrule
    K-Means & 79.78 \\
    LSTM & 49.45 \\
    Transformer & 78.22 \\
    Character Query Transformer & \textbf{92.53} \\
    \bottomrule
  \end{tabular}
  \caption*{mean Intersection over Union (mIoU)}
\end{table}

\subsubsection{Combined}

Finally, the models have been trained using the combined training sets, in order to see whether they are capable of
scaling to multiple languages and improve the overall results by attaining additional information that can be found in
the other dataset. It has to be noted that because these models use embeddings of the characters, the mutually exclusive
characters are not directly benefiting from the combination of the dataset, in the sense of having more data points of
the same character, but can still improve as the model's general segmentation capability improves.
Even though $k$-means is not affected by changing the training data, it is still listed in \autoref{tab:results} alongside
the others for reference. The deterioration of the LSTM on the IAM-OnDB was foreseeable as it was not able to properly
learn from the HANDS-VNOnDB. The drop of 6.48\% (from 93.72\% to 87.24\%) on the \textit{Test Set F} is significant but not to
the point where the model fails completely. At the same time, its results on the HANDS-VNOnDB improved a little, from
49.45\% to 53.66\% on the \textit{Test Set}, but less than the IAM-OnDB degraded.
When it comes to the Transformer with the character boundary prediction, it is almost identical on the HANDS-VNOnDB as it was without using both datasets to train, but similarly to the LSTM the result on the IAM-OnDB \textit{Test Set F} deteriorated from 90.34\% to 86.18\% (-4.16\%), indicating that combining the two datasets has a negative effect on the models predicting character boundaries.
The Character Query Transformer is the only model that benefited from training on both datasets. Even though the results
on the HANDS-VNOnDB barely changed (-0.47\%), the IAM-OnDb improved by almost 8\% (from 87.48\% to 95.11\%).
This demonstrates that the character queries are robust and that it is capable of scaling to multiple languages, especially
as the additional data contributed to the large data requirements of the Transformer, even though it was not data from the same language.

\begin{table}[ht]
  \centering
  \caption{
    \textbf{Combined datasets results.} All models were trained using the combined training sets
    of IAM-OnDB and HANDS-VNOnDB and the best model was determined by the mIoU across the validation sets.
  }\label{tab:results}
  \begin{tabular}{c|cc|c}
    \toprule
    & \multicolumn{2}{c|}{IAM-OnDB} & HANDS-VNOnDB \\
    Model & \textit{Test Set T} & \textit{Test Set F} & \textit{Test Set} \\
    \midrule
    K-Means & 88.94 & 91.05 & 79.78 \\
    LSTM & 82.70 & 87.24 & 53.66 \\
    Transformer & 80.93 & 86.18 & 78.25 \\
    Character Query Transformer & \textbf{92.28} & \textbf{95.11} & \textbf{92.06} \\
    \bottomrule
  \end{tabular}
  \caption*{mean Intersection over Union (mIoU)}
\end{table}

\subsection{Ablation Study: CTC Spikes}

CTC spikes can be used as additional information whenever a CTC-based recognizer has been run beforehand, as it already broadly
located the characters and therefore can serve as an initial guidance. There are other cases, where either the text was
already known without having to run a recognizer, or when the recognizer does not utilize CTC. In this ablation study we
remove the CTC spikes to see whether they are a meaningful addition to the models. In the case of $k$-means, the points
where the CTC spikes occurred were used as the initial centroids, without the CTC spikes they are randomly initialized
instead, as is common practice. All other models simply do not have the CTC spike information in the points.
The ablation was conducted on the combined datasets.

Including the CTC spikes is a significant improvement across the board. The difference between using the CTC spikes and
not, varies depending on the model, ranging from 1.89\% for the Transformer on the HANDS-VNOnDB \textit{Test Set}
up to 11.88\% for the character boundary predicting Transformer on the IAM-OnDB \textit{Test Set F}. Generally, the
CTC spikes are less impactful on the HANDS-VNOnDB, with the exception for the LSTM. The Character Query Transformer is the
most consistent and hovers around a difference of 4.5\% on all datasets, suggesting that it is very stable and is
not tied to the CTC spikes but simply uses them to improve the results in a meaningful way. Even though the results
without the CTC spikes are not quite as good, they can still be used in cases where no CTC spikes are available.

\begin{table}[ht]
  \centering
  \caption{
    \textbf{CTC spikes ablation study.}
    Comparison of results when using CTC spikes as a feature and without it. Including the CTC spikes improves the
    results significantly across all models.
  }\label{tab:results-ctc}
  \begin{tabular}{cc|cc|c}
    \toprule
     & & \multicolumn{2}{c|}{IAM-OnDB} & HANDS-VNOnDB \\
    Model & CTC Spikes & \textit{Test Set T} & \textit{Test Set F} & \textit{Test Set} \\
    \midrule
    K-Means & & 80.12 & 83.83 & 76.82 \\
    K-Means & \checkmark & 88.94 & 91.05 & 79.78 \\
    \hline
    LSTM & & 74.60 & 80.59 & 42.24 \\
    LSTM & \checkmark & 82.70 & 87.24 & 53.66 \\
    \hline
    Transformer & & 70.21 & 74.30 & 76.36 \\
    Transformer & \checkmark & 80.93 & 86.18 & 78.25 \\
    \hline
    Character Query Transformer & & 86.03 & 90.78 & 87.58 \\
    Character Query Transformer & \checkmark & \textbf{92.28} & \textbf{95.11} & \textbf{92.06} \\
    \bottomrule
  \end{tabular}
  \caption*{mean Intersection over Union (mIoU)}
\end{table}

\section{Conclusion}

In this paper, we have introduced a novel Transformer-based approach to on-line handwritten character segmentation,
which uses learned character queries in the Transformer decoder block to assign sampling points of stylus trajectories to the
characters of a known transcription. In an experimental evaluation on two challenging datasets, IAM-OnDB and
HANDS-VNOnDB, we compare the proposed method with $k$-means, LSTM, and a standard Transformer architecture. In comparing
the four methods, we observe that approaches which rely on spatial information ($k$-means) perform reasonably well on
non-monotonic handwriting but lack learned features to extract the exact character boundaries. The approaches that rely
on temporal information (LSTM and standard Transformer) perform well on mostly-monotonic handwriting, but fail in highly
non-monotonic cases. Using the Transformer decoder block in combination with character queries allows us to outperform all
other approaches because it uses the strengths of the learned solutions, but does not have a strong inductive bias
towards monotonic handwriting.

We provide a character segmentation ground truth for the IAM-OnDB and HANDS-VNOnDB using a high-quality approximation.
Producing a perfect ground truth for on-line handwritten character segmentation is impossible for cursive script, since
even humans will not always agree on the exact start and end positions of the characters. Therefore, in future work, we
aim to encode this uncertainty into the ground truth and into the evaluation measures. Another line of future research
is to use the segmented characters for creating additional synthetic training material, which is expected to further
improve the performance of the Character Query Transformer.

\subsection*{Acknowledgement}
We would like to thank the following contributors for ideas, discussions, proof-reading, and support:  Henry A. Rowley and Pedro Gonnet. 

\printbibliography

\end{document}